\documentclass[10pt,twocolumn,letterpaper]{article}

\usepackage[pagenumbers]{cvpr} 

\usepackage{multirow}
\usepackage{tabularx}
\usepackage{rotating}
\usepackage{booktabs}
\usepackage{amssymb}
\usepackage{array}
\usepackage{threeparttable}
\usepackage{multirow}
\usepackage{adjustbox}
\usepackage{bm}
\usepackage{array}
\usepackage{wrapfig}

\newcolumntype{C}[1]{>{\centering\arraybackslash}p{#1}}

\newcommand{\smallleq}{\text{\tiny$\leq$}}

\newcommand{\ospaps}{$\mathcal{O}_{PS}$}

\newcommand{\ospapsbold}{$\bm{\mathcal{O}}_{\textbf{PS}}$}
\newcommand{\ospapsboldt}{$\bm{\mathcal{O}}^{\textbf{Th}}_{\textbf{PS}}$}
\newcommand{\ospapsbolds}{$\bm{\mathcal{O}}^{\textbf{St}}_{\textbf{PS}}$}

\newcommand{\ospapt}{$\mathcal{O}^2_{PT}$}

\newcommand{\ospapts}{$\mathcal{O}^{2S}_{PT}$}

\newcommand{\ospaptbold}{$\bm{\mathcal{O}}^{\textbf{2}}_{\textbf{PT}}$}
\newcommand{\ospaptboldl}{$\bm{\mathcal{O}}^{\textbf{2}}_{\textbf{\textit{Loc}}}$}
\newcommand{\ospaptboldc}{$\bm{\mathcal{O}}^{\textbf{2}}_{\textbf{\textit{Card}}}$}
\newcommand{\ospaptboldk}{$\bm{\mathcal{O}}^{\textbf{2K}}_{\textbf{PT}}$}
\newcommand{\ospaptboldu}{$\bm{\mathcal{O}}^{\textbf{2U}}_{\textbf{PT}}$}
\newcommand{\ospaptboldt}{$\bm{\mathcal{O}}^{\textbf{2T}}_{\textbf{PT}}$}
\newcommand{\ospaptbolds}{$\bm{\mathcal{O}}^{\textbf{2S}}_{\textbf{PT}}$}

\newcommand{\smaller}[1]{{\tiny\textbf{#1}}} 

\usepackage[dvipsnames]{xcolor}

\definecolor{cvprblue}{rgb}{0.21,0.49,0.74}
\usepackage[pagebackref,breaklinks,colorlinks,citecolor=cvprblue]{hyperref}

\def\paperID{8798} 
\def\confName{CVPR}
\def\confYear{2024}

\title{JRDB-PanoTrack: An Open-world Panoptic Segmentation and Tracking Robotic Dataset in Crowded Human Environments}
\author{Duy Tho Le$^{1}$\footnotemark[1] , Chenhui Gou$^{1}$\footnotemark[1]\ , Stavya Datta$^{1}$, Hengcan Shi$^{1}$\footnotemark[2], \\ Ian Reid$^{2,3}$, Jianfei Cai$^{1}$, Hamid Rezatofighi$^{1}$\\
$^{1}$Monash University, $^{2}$MBZUAI, $^{3}$University of Adelaide, \\
{\tt\small\{tho.le1, chenhui.gou, hengcan.shi\}@monash.edu} \\
{\tt\small\footnotemark[1] Equal contribution, \footnotemark[2] Corresponding author} \\
\small\url{https://jrdb.erc.monash.edu/dataset/panotrack}
}

\begin{document}
\maketitle
\begin{abstract}
Autonomous robot systems have attracted increasing research attention in recent years, where environment understanding is a crucial step for robot navigation, human-robot interaction, and decision. Real-world robot systems usually collect visual data from multiple sensors and are required to recognize numerous objects and their movements in complex human-crowded settings. Traditional benchmarks, with their reliance on single sensors and limited object classes and scenarios, fail to provide the comprehensive environmental understanding robots need for accurate navigation, interaction, and decision-making. As an extension of JRDB dataset, we unveil JRDB-PanoTrack, a novel open-world panoptic segmentation and tracking benchmark, towards more comprehensive environmental perception. JRDB-PanoTrack includes \textbf{(1)} various data involving indoor and outdoor crowded scenes, as well as comprehensive 2D and 3D synchronized data modalities; \textbf{(2)} high-quality 2D spatial panoptic segmentation and temporal tracking annotations, with additional 3D label projections for further spatial understanding;    \textbf{(3)} diverse object classes for closed- and open-world recognition benchmarks, with OSPA-based metrics for evaluation. Extensive evaluation of leading methods shows significant challenges posed by our dataset.

\end{abstract}    

    \begin{table*}
        \centering
        \footnotesize
        \resizebox{\textwidth}{!}{
            \begin{tabular}{C{2.5cm}|C{0.5cm}C{0.5cm}C{0.5cm}|C{0.5cm}C{0.5cm}C{0.9cm}|C{0.5cm}C{0.5cm}|C{0.65cm}|C{0.5cm}C{0.7cm}C{0.9cm}C{0.7cm}}
                \toprule
                & \multicolumn{3}{c|}{\textbf{Data}} & \multicolumn{3}{c|}{\textbf{Domain}} & \multicolumn{2}{c|}{\textbf{ No. Class}} & & & & & \\
                \cmidrule{1-14}
                \textbf{Dataset} &
                \textbf{Data} &
                \textbf{Temp} &
                \textbf{Pano Cov.} &
                \textbf{In door} &
                \textbf{Out door} &
                \textbf{Platform} &
                \textbf{Thing} &
                \textbf{Stuff} &
                \textbf{Open world} &
                \textbf{Trk Len} &
                \textbf{No. Seq} &
                \textbf{No. Smp} &
                \textbf{No. M} \\
                \cmidrule{1-14}
                \textbf{PanopticCOCO} \cite{coco}           & I   &            &      & \checkmark & \checkmark &     Int      & 80 & 91 &            & -             &      & 164k &      \\
                \textbf{Cityscapes} \cite{cityscapes}       & I   & \checkmark &      & \checkmark &            &     Car       & 8  & 11 &            &               & 500  & 3k   & 10k  \\
                \textbf{VIPSeg} \cite{vipseg}               & I   & \checkmark &      & \checkmark & \checkmark &     Int      & 58 & 66 &            & \smallleq10s  & 3536 & 85k  & 926k \\
                \textbf{MOT-STEP}  \cite{step}                & I   & \checkmark &      &            & \checkmark &   Int         & 1  & 6  &            & \smallleq19s  & 4    & 2k   & 17k  \\
                \textbf{KITTI-STEP}  \cite{step}              & I   & \checkmark &      &            & \checkmark &   Car         & 2  & 17 &            & \smallleq65s  & 50   & 19k  & 126k \\
                \textbf{Waymo }  \cite{waymo}                  & I   & \checkmark & 220° &            & \checkmark &  Car          & 8  & 20 &            & \smallleq1.2s & 2060 & 100k &      \\
                \textbf{SemanticKITTI} \cite{semantickitti} & P   & \checkmark &      &            & \checkmark &     Car       & 14 & 11 &            &               & 21   & 43k  &      \\
                \textbf{Nuscenes}  \cite{nuscenes}            & P   & \checkmark & 360° &            & \checkmark &   Car         & 23 & 6  &            &               & 1000 & 40k  & 1.2M \\
                \cmidrule{1-14}
                \textbf{JRDB-PanoTrack }                    & I/P & \checkmark & 360° & \checkmark & \checkmark & Rob & 60 & 11 & \checkmark & \smallleq117s & 54 & ~20k & 428k \\
                \bottomrule
            \end{tabular}}
        \caption{Typical datasets for 2D-3D panoptic segmentation and tracking. Abbreviations: I (Image), P (Point Cloud), Car (Autonomous Car), Rob (Mobile Robot), Int (Internet images/videos), Temp (Temporal data), Pano Cov. (Panoramic Coverage), No. Class (The number of classes), 
            Trk Len (Track Length), No. Seq (The number of sequences), No. Smp (The number of samples) and No. M (the number of masks).}
        \label{tab:dataset_comparison}
    \end{table*}


\section{Introduction}
%


With the increasing demands for autonomous robots in human-crowded environments, environment understanding becomes paramount, which serves as a vital step in many robotic systems, such as navigation and human-robot interaction. Specifically, human-centric environment understanding can be mainly divided into two aspects: spatial and temporal understanding. Spatial understanding aims to distinguish objects in human-crowded environments, while temporal understanding expects to recognize temporal relations of such objects.


The existing datasets for environment understanding, sourced primarily from self-driving vehicles \cite{cityscapes, semantickitti, waymo, nuscenes}, or internet images/videos \cite{coco, vipseg}, exhibit clear domain gaps when applied to robotic environments. These sources typically offer different perspectives from robots, and fail to encapsulate the challenges and interactions specific to robotic systems. As shown in \cref{tab:dataset_comparison}, most of the existing datasets only contain a single data modality (RGB images or point clouds) \cite{cityscapes, vipseg} and a small number of classes \cite{step, cityscapes}. They also lack temporal information \cite{coco} or 360-degree panoramic spatial perspectives \cite{cityscapes, step, waymo}. In contrast, real-world applications where the robotic agents are deployed, usually involve multi-modal data, diverse classes, and both spatial and temporal understanding. 

Built on top of JRDB dataset \cite{jrdb, jrdbact, jrdbpose}, inheriting its comprehensive annotation suite for human bodies, we introduce \textit{JRDB-PanoTrack}, a novel comprehensive dataset for human-crowded environment understanding. 
Firstly, JRDB-PanoTrack offers a comprehensive dataset from various indoor and outdoor crowded scenes with 2D and 3D synchronized data modalities, supporting visual and robotic applications. Secondly, high-quality 2D panoptic segmentation and tracking annotations are provided for both spatial and temporal environment understanding, including 428K panoptic masks, 27K tracking labels and 7.3B annotated pixels. Additional 3D label projections are also presented for further spatial understanding. Thirdly, we introduce diverse objects and open-world benchmarks for generalization research.   
Finally, JRDB-PanoTrack annotates multiple classes for some areas, such as objects behind \textit{glass} or hang on \textit{wall} in \cref{fig:thumbnail}. We propose metrics based on optimal sub-pattern matching (OSPA) to deal with such evaluation. 



Based on the JRDB-PanoTrack dataset, we present several benchmarks, including \textit{Closed-world (CW)} and \textit{Open-world (OW)} panoptic segmentation and tracking. We extensively evaluate state-of-the-art (SOTA) methods on these benchmarks. Moreover, SOTA methods are also estimated on our 3D label projections. 
The results underline the imperative need for advanced methodologies that can adeptly handle the complexities presented by complex human-crowded environments. Our main contributions are:
\begin{itemize}
    \item We present JRDB-PanoTrack, an extensive new dataset for spatial and temporal robotic environment understanding. In JRDB-PanoTrack, high-quality panoptic segmentation and tracking annotations are provided. 
    We employ comprehensive data collected by a mobile robot, including 2D\&3D modalities as well as indoor\&outdoor human-crowded scenes.
    \item Closed- and open-world benchmarks are proposed for generalizable environment understanding. Our dataset also contains multi-class annotations and OSPA-based metrics for evaluation.
    \item We conduct extensive evaluations of SOTA closed- and open-world segmentation/tracking methods on JRDB-PanoTrack, and discuss their strengths and weaknesses.
\end{itemize}




\section{Related Work}

\begin{figure*}[!t]
    \centering
    \includegraphics[width=\linewidth]{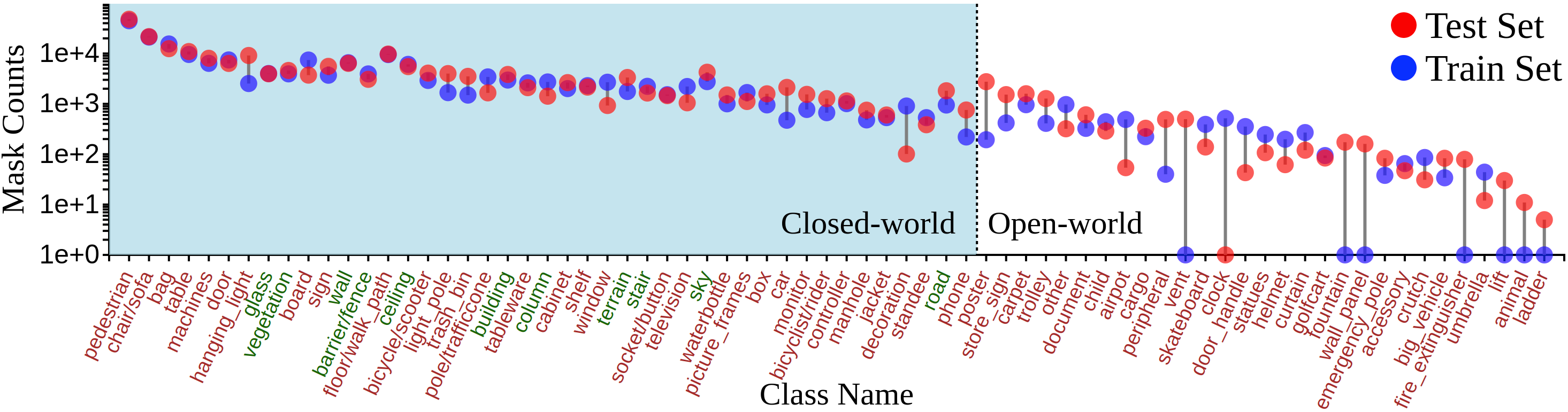}
    \caption{ Distribution of object masks of \textit{thing} (brown) and \textit{stuff} (green) classes in JRDB-PanoTrack train and test sets, where x and y-axis indicate the class names and mask counts, respectively. Best viewed in color and zoomed in.}
    \label{fig:instances_lolipop}
\end{figure*}

\textbf{Panoptic Segmentation and Tracking Datasets.} \textit{\textbf{Panoptic segmentation}}, as introduced in \cite{panopticsegmentation}, is a task to generate instance-level masks for \textit{thing} objects (countable, distinct entities) and class-level masks for \textit{stuff} objects (amorphous and uncountable regions) to achieve a more complete visual understanding. Datasets like PanopticCOCO \cite{coco}, ADE20K \cite{ade20k} and Cityscapes \cite{cityscapes} are widely popular in this space, primarily focusing on 2D images. However, these datasets only support spatial understanding. 

\noindent\textit{\textbf{Panoptic tracking}} further integrates multi-object tracking into panoptic segmentation, as seen in datasets like MOT-STEP \cite{step}, VIPSeg \cite{vipseg}, and Waymo \cite{waymo} for 2D tracking, and SemanticKITTI \cite{semantickitti}, Panoptic Nuscenes \cite{nuscenes} for 3D tracking. However, these datasets, often sourced from self-driving cars \cite{nuscenes, waymo, vps, step, semantickitti}, single-view surveillance cameras \cite{step} or miscellaneous internet videos \cite{vipseg}. These datasets, although useful and large-scale, fall short in representing complex, human-centric environments for autonomous robotics due to the lack of synchronized multi-modal multi-view data, diverse object classes, complex human-crowded scenes, and domain consistency. Our JRDB-PanoTrack dataset addresses this gap by providing synchronized 2D and 3D data from a social mobile manipulator, capturing the complexity of crowded human spaces, offering diversity in objects and unique challenges in both closed-world and open-world settings.


\noindent\textbf{OW Benchmarks.} The development of OW benchmarks is crucial for assessing the generalization capabilities of models in diverse and unpredictable environments. Large-scale segmentation datasets such as COCO \cite{coco} and ADE20K \cite{ade20k} and OW segmentation datasets \cite{peng2023openscene} are usually used for OW spatial understanding, while several datasets like TAO-OW \cite{taoow} and OVTrack \cite{li2023ovtrack} are introduced for OW bounding box tracking. Moreover, these datasets are all from internet images/videos. Different from them, JRDB-PanoTrack introduces a unique and challenging OW benchmark for panoptic segmentation and tracking in robotic environments, with both 2D and 3D data modalities.

\noindent\textbf{Previous JRDB Datasets.} JRDB \cite{jrdb} is a large-scale and comprehensive dataset for autonomous robot research in human-centric environments. It collects 2D and 3D point cloud videos, audio as well as GPS positions by a social manipulator robot. In previous JRDB \cite{jrdb}, JRDB-Act \cite{jrdbact} and JRDB-Pose \cite{jrdbpose}, 2D-3D human detection, tracking and forecasting, body skeleton pose estimation, human social grouping and activity recognition annotations have been introduced. In JRDB-PanoTrack, we complement this JRDB by providing new open-world panoptic segmentation and tracking annotations for more comprehensive human-centered scene understanding.

\noindent\textbf{SOTA Frameworks.} 
For \textbf{\textit{panoptic segmentation}}, initial approaches \cite{panopticsegmentation, kirillov2019panoptic2, xiong2019upsnet} handle semantic and instance segmentation as separate tasks by using dual sub-networks. Max-deeplab \cite{wang2021max} introduces transformer-based architectures, moving away from bounding box-dependent models. Recent developments, including K-net \cite{zhang2021k}, MaskFormer \cite{cheng2021per}, Mask2Former \cite{cheng2022masked} and Mask DINO \cite{li2023mask}, unifies semantic, instance and panoptic segmentation into a singular mask proposal prediction framework. In the OW domain, methods like \cite{ding2022open, xu2023open, qin2023freeseg, xu2023masqclip} generate mask proposals for all panoptic objects, and then align them with object names via large vision-language pre-training. 

\noindent For \textbf{\textit{multi-object tracking}}, traditional motion-model-based algorithms often outperform modern integrated systems. SORT \cite{sort} exemplifies this with its linear-motion-based track association. ByteTrack \cite{bytetrack} introduces low-confidence detection associations to improve tracking. OC-SORT \cite{ocsort} enhances filters and recovery strategies to solve the non-linear motion problem. More recently, BoT-SORT \cite{botsort} advances the field by optimizing the Kalman filter state and incorporating camera-motion compensation. Notably, to the best of our knowledge, there is no available OW panoptic tracking method. We use those strong and popular trackers as baselines in our experiments.

\section{The JRDB-PanoTrack Dataset}
\subsection{Dataset and Statistics}


\textbf{Data.} JRDB-PanoTrack encompasses 20,000 images, sampled at 1Hz from 54 videos in the original JRDB dataset \cite{jrdb}. 4,000 360-degree panoramic images can be generated by merging 5 original images from 5 different camera views. 4,000 point clouds are also provided for 3D understanding. 

\noindent\textbf{Annotation.} JRDB-PanoTrack retains all annotations from JRDB\cite{jrdb} and further enhances the dataset by introducing 428K 2D panoptic segmentation and 27K tracking annotations to enable environment understanding.

\noindent\textbf{Annotation process.} The annotation process starts with an unlimited list of classes which can be extended by all annotators, any clearly visible and semantically meaningful objects would be annotated, objects that are behind the glass or being hang on wall will be annotated with multiple labels. Then annotators produce labels and senior annotators control the quality by multiple inspection rounds.

\noindent\textbf{Object Class.} There are 72 objects in JRDB-PanoTrack, which are classified into 61 \textit{thing} (such as pedestrians, cars and laptops) and 11 \textit{stuff} (like sky and walls) classes. \cref{fig:instances_lolipop} depicts the distributions of classes.

\begin{figure}[!t]
    \centering
    \includegraphics[width=\linewidth]{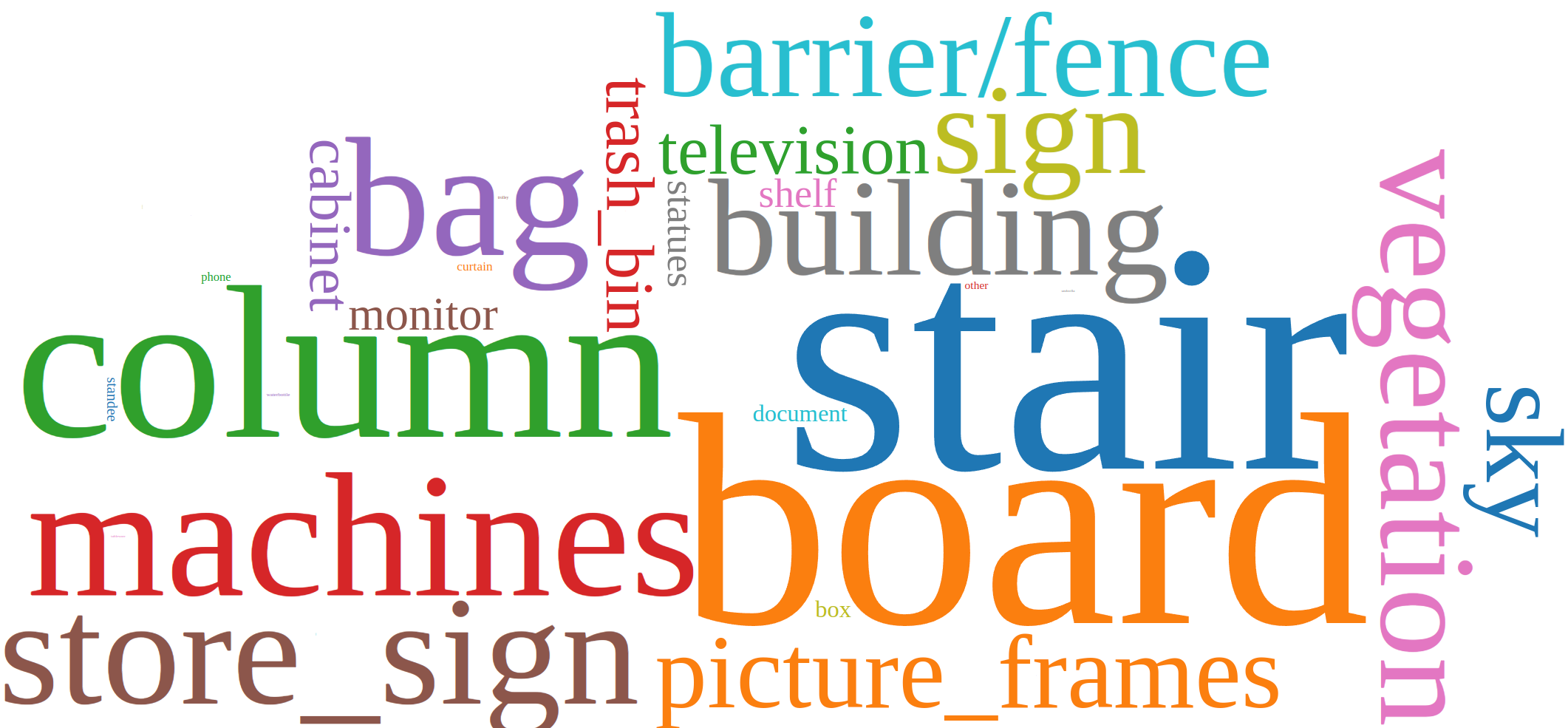}
    \caption{Word cloud of the most frequent classes seen through glass in JRDB-PanoTrack, with the size of the word proportional to the frequency of the class.}
    \label{fig:behind_glass_wordcloud}
    \vspace{-1em}
\end{figure}

\noindent\textbf{Special Class Labeling.} Our dataset aims to analyze common environments for autonomous robots. There are some differences from traditional environment understanding datasets. \textbf{(1) \textit{Floor differentiation}}: human-robot interaction and navigation require robots to distinguish different floors. To address this, we provide instance segmentation labels for floors and regard them as \textit{thing} objects. \textbf{(2) \textit{Multi-class segmentation}}: In modern environments, objects are often seen behind \textit{windows} and \textit{glass}, and sometimes being hang on walls (most frequently seen objects are shown in \cref{fig:behind_glass_wordcloud}). Traditional datasets usually simply ignore these objects or interaction, while they might be crucial for environment understanding. In JRDB-PanoTrack, there are 9\% of objects belonging to such cases. Therefore, we label multiple classes for pixels belonging to these objects, \ie, including the front \textit{windows} or \textit{glass}, and the behind objects. We hope this will encourage the community to develop more robust models for better scene understanding.

\begin{figure}[t]
    \centering
    \includegraphics[width=\linewidth]{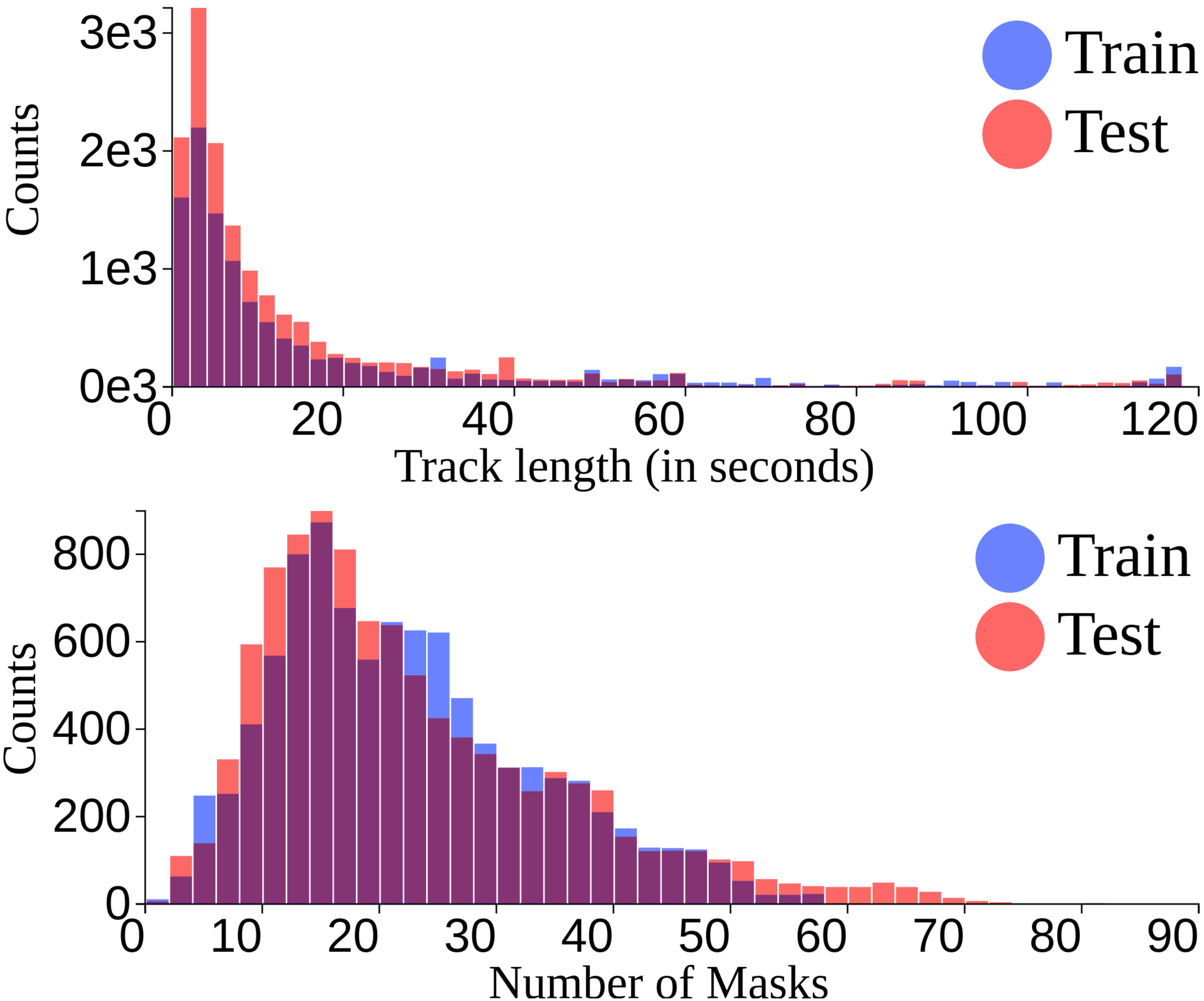}
    \caption{Analysis of Track length distribution (top) and Number of masks per frame (bottom) in the JRDB-PanoTrack dataset. Best viewed in color.}
    \label{fig:track_inst_stats}
\end{figure}

\noindent\textbf{Tasks.} Our dataset supports panoptic segmentation and tracking tasks. \textbf{\textit{Panoptic segmentation}} \cite{panopticsegmentation} expects to spatially understand environments, which generates masks for all the \textit{thing} and \textit{stuff} objects. \textbf{\textit{Panoptic tracking}} \cite{Mopt} understands environments on both spatial and temporal aspects. It segments both \textit{thing} and \textit{stuff} objects, and tracks \textit{thing} objects throughout a video. 
As shown in \cref{fig:track_inst_stats}, there are up to 81 masks in an image, and the average number of masks is 22. In panoramic views, the maximum and average mask counts per image are 245 and 80, respectively.  
\cref{fig:track_inst_stats} also highlights the track length distribution in JRDB-PanoTrack. The maximum and average track lengths are 117s and 16s, respectively. The most populated scene in our dataset comprises a staggering 564 tracklets (1010 in panoramic views) in a single sequence, compared to the average 101 tracklets (198 in panoramic views) per sequence. According to \cref{fig:track_inst_stats}, the testing set is more crowded than the training set with more masks per image and more tracklets per sequence.

\begin{table}[!tp]
\vspace{-1em}
    \centering
    \begin{adjustbox}{width=\columnwidth,center}
    \small
    \begin{tabular}{C{7mm}|C{6mm}C{6mm}C{6mm}C{6mm}C{6mm}C{6mm}C{6mm}}
        \toprule
        & \multicolumn{2}{c}{\textbf{\# Inst. / img}} & \multicolumn{2}{c}{\textbf{\# Trk. / seq.}} & \multicolumn{2}{c}{\textbf{Track length}} \\
        \cmidrule(lr){2-3} \cmidrule(lr){4-5} \cmidrule(lr){6-7}
        & \textbf{Mean} & \textbf{Max} & \textbf{Mean} & \textbf{Max} & \textbf{Mean} & \textbf{Max} \\
        \midrule
        \textbf{Train} & 22 / 78*      & 57 / 144*    & 87 / 178*     & 249/ 420*    & 17 / 27*      & 116/ 116*    \\
        \textbf{Test}  & 22 / 80*      & 81 / 245*    & 105/ 217*     & 564/ 1010*   & 14 / 24*      & 117/ 117*    \\
        \bottomrule
    \end{tabular}
    \end{adjustbox}
    \footnotesize{[*] statistics for panoramic images}
    \vspace{-1em}
    \caption{The number of masks per image, the number of tracklets per sequence, as well as track lengths (in seconds).}
    \label{tab:track_inst_stats}
    \vspace{-1.2em}
\end{table}

\noindent\textbf{Thing and Stuff classes.} Following \cite{panopticsegmentation}, we divide the object classes into \textit{thing} and \textit{stuff} classes. \textit{Thing} classes are objects that can be segmented and tracked, such as \textit{person, car, bicycle, chair, table, laptop, bottle, etc.} \textit{Stuff} classes are background classes that can be segmented but not tracked, such as \textit{sky, ground, wall, etc.} \cref{fig:instances_lolipop} shows the distribution of object instances in JRDB-PanoTrack, where \textit{pedestrian} is the most frequent class with more than 40k instances, followed by the commonly seen objects in human-centric environments such as \textit{chair, bag, table, door, board, machines, etc.} with 10k to 50k instances. 

\noindent\textbf{CW and OW.} Based on class distributions shown in \cref{fig:instances_lolipop}, we divide our 72 panoptic classes into two sets: 43 common and 29 long-tail classes as \textit{known} and \textit{unknown} classes, respectively. The 43 \textit{known} classes can be used for training and evaluation at the closed-world (\textbf{\textit{CW}}) scenario. At the open-world (\textbf{\textit{OW}}) scenario, the 43 \textit{known} classes can be employed for training, while 28 \textit{unknown} classes are for testing (there is one class that occur in training set only).

\noindent\textbf{Tracklet statistics.} \cref{tab:track_inst_stats} offers detailed statistics on the number of masks per image, tracklets per sequence, and track lengths in the JRDB-PanoTrack dataset. It underscores the dataset's depth and diversity, with some tracks extending up to 117 seconds across multiple camera views. According to \cref{tab:track_inst_stats}, the testing set is more crowded than the training set with more masks per image and more tracklets per sequence.

\begin{figure}[t]
     \centering
     \includegraphics[width=\linewidth]{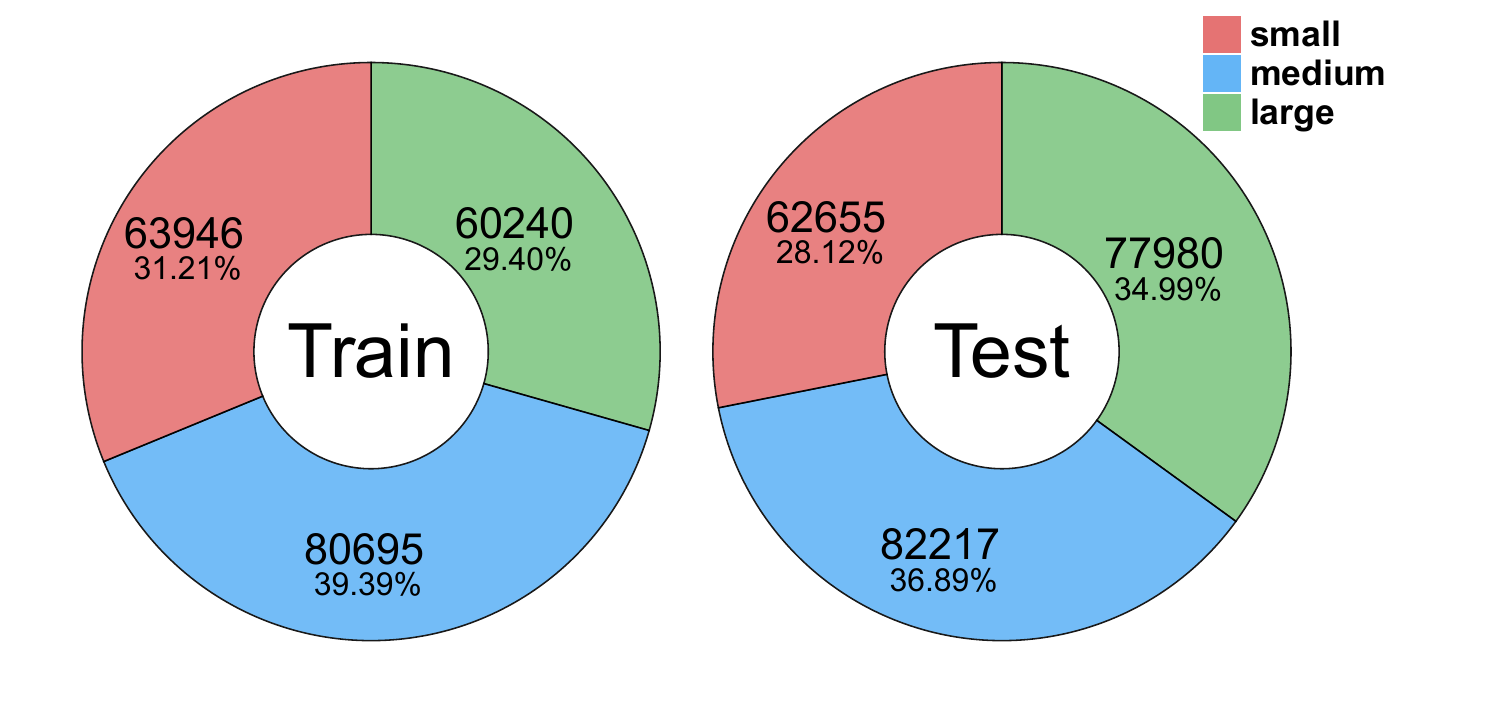}
     \caption{The count (top) and percentage (bottom) of Small, Medium and Large masks in JRDB-PanoTrack training and testing sets. Small and Large masks are the masks $\leq 32^2$ and $\leq 96^2$ pixels, and the sizes of Medium masks are in between. Image size is 752x480 (W x H).}
     \label{fig:mask_size_per_split}
 \end{figure}

\noindent\textbf{Mask Size.} The distribution of mask sizes in JRDB-PanoTrack are as presented in \cref{fig:mask_size_per_split}. We have balanced mask sizes in both training and testing sets, which bring challenges to panoptic segmentation and tracking methods to carefully deal with objects of various sizes.

\subsection{Benchmark and Metrics}
\textbf{Benchmark.} Based on our JRDB-PanoTrack dataset, we propose several benchmarks for environment understanding, the four categories are:
\begin{itemize}
    \item \textbf{\textit{CW panoptic segmentation.}}
    \item \textbf{\textit{OW panoptic segmentation.}}
    \item \textbf{\textit{CW panoptic tracking.}}
    \item \textbf{\textit{OW panoptic tracking.}}
\end{itemize}
In all benchmarks, we use half of our dataset for training, i.e., 9365 images in 27 sequences. For testing, we employ 9280 images in the other 27 sequences. Panoramic images and point clouds corresponding to the 9365/9280 images can be used for panorama and 3D understanding. In CW benchmarks, we release annotations of the 43 \textit{known} classes for both training and testing. In OW benchmarks, methods can use \textit{known} classes for training, while being tested on all of the classes.

\vspace{5pt}

\noindent\textbf{Metric.} Current evaluation methods for panoptic segmentation, despite their utility, exhibit limitations that can skew method rankings, due to:
\begin{itemize}
\item Threshold-based, where the choice of the threshold can change the ranking of the methods, making it unreliable \cite{howtrustworthy}: VPQ~\cite{vps} and PTQ~\cite{Mopt}.
    \item Inadvertently penalizing the rectification of errors (ID recovery): VPQ~\cite{vps} and PTQ~\cite{Mopt}.
    \item Inability to handle multi-label scenarios: VPQ~\cite{vps}, PTQ~\cite{Mopt}, and STQ~\cite{step}.
\end{itemize}
Given the introduction of multi-label panoptic segmentation and tracking by JRDB-PanoTrack, these existing metrics become insufficient. To address the gaps, we introduce OSPA$_{PS}$ and OSPA$^2_{PT}$, specifically designed for panoptic segmentation and tracking.

\begin{table}[!tb]\centering
\scriptsize
\begin{adjustbox}{width=\linewidth}
    \begin{tabular}{p{1.1cm}|C{0.35cm}C{0.4cm}|C{0.35cm}C{0.43cm}|C{0.35cm}C{0.43cm}}
        \toprule
        \textbf{Method} & \textbf{PQ$\uparrow$}  & \textbf{\ospapsbold$\downarrow$}   & \textbf{PQ$^{\textbf{Th}}$$\uparrow$} & \textbf{\ospapsboldt$\downarrow$}  & \textbf{PQ$^{\textbf{St}}$$\uparrow$} &\textbf{\ospapsbolds $\downarrow$}\\
        \midrule
       \textbf{kMaX}\cite{yu2022k}             & 32.52  & 0.67 & 27.96&0.72 & 45.81& 0.53\\
        \textbf{2Former}\cite{cheng2021mask2former} & 33.25   &0.66 & 28.74& 0.71&46.38 &0.52 \\
        \textbf{DINO}\cite{li2023mask}             & \textbf{36.57}&\textbf{0.64} &\textbf{33.07} &\textbf{0.68} &\textbf{46.74} & \textbf{0.51}\\
        \bottomrule
    \end{tabular}
\end{adjustbox}
\caption{\small{\textbf{Results of CW panoptic segmentation methods on JRDB-PanoTrack.} All methods use ResNet-50 backbone and COCO pre-training. There are 43 classes, 32 \textit{Thing} and 11 \textit{Stuff}. (Kmax for Kmax-Deeplab, 2Former for Mask2Former and DINO for mask DINO.)}}
\label{tab:panoptic:closeworld_16th_11}
\end{table}
\noindent\textbf{OSPA for Panoptic Segmentation.} The Optimal Sub-Pattern Matching (OSPA) metric, known for incorporating miss-distance in multi-object performance evaluation \cite{ospa}, has recently been adapted for bounding box/pose detection and tracking tasks \cite{howtrustworthy, jrdbpose}. 
Building on this, we introduce OSPA$_{PS}$ (\ospaps), a variant of OSPA, specifically designed for multi-label panoptic segmentation.

Let $X = \{x_1, x_2, \dots, x_m\}$ and $Y = \{y_1, y_2, \dots, y_n\}$ be two sets of arbitrary mask regions ($x,y\subset \mathbb{R}^2$) on an image for all ground-truths and predictions, with cardinalities $|X|$ and $|Y|$, where $|Y| \geq |X|$ (otherwise flip $X, Y$). For a given class $c \in \mathbb{C}$, we calculate the normalised base distance between masks $d_{K}(x_i, y_i) = 1 - IOU(x_i, y_i) \in [0, 1]$. $\mathcal{O}_{PS}(X_c, Y_c)$ is then acquired by using OSPA equation \cite{ospa}. The overall OSPA error is calculated by averaging the OSPA error over all classes:
\begin{equation}
    \mathcal{O}_{PS}(X, Y) = \frac{1}{|\mathbb{C}|}\sum_{c \in \mathbb{C}}\mathcal{O}_{PS}(X_c, Y_c).
    \label{eq:ops}
\end{equation}

\noindent\textbf{OSPA for Panoptic Tracking.}  The OSPA metric is expanded to assess panoptic tracking with the introduction of $OSPA_{PT}$ (\ospapt). For each class $c \in \mathbb{C}$, consider $\mathbf{X}_c = \{X^{\mathcal{D}1}_{1c}, X^{\mathcal{D}2}_{2c}, \ldots, X^{\mathcal{D}m}_{mc}\}$ and $\mathbf{Y}_c = \{Y^{\mathcal{D}1}_{1c}, Y^{\mathcal{D}2}_{2c}, \ldots, Y^{\mathcal{D}n}_{nc}\}$ as sets of mask trajectories for ground-truth and predicted masks, respectively, where $\mathcal{D}i$ contains the time indices where track $i$ exists. Then, we calculate the time average distance of every pair of tracks $X^{\mathcal{D}_i}_{ic}$ and $Y^{\mathcal{D}_j}_{jc}$ similar to \cite{ospa} using OSPA set distance $d_{O}(\{X_{ic}^{t}\},\{Y_{jc}^{t}\} = 1 - IOU(x^t_{ic}, y^t_{jc})$. If only $\{X_i^{t}\}$ or $\{Y_j^{t}\}$ exists,  then $d_{O}(\{X_{ic}^{t}\},\{Y_{jc}^{t}\}) = 1$, otherwise $d_{O}(\{X_{ic}^{t}\},\{Y_{jc}^{t}\})=0$. The remaining step remained the same as the original OSPA$^2$, \ospapt\ is the average of all classes similar to \cref{eq:ops}.

\begin{table}[t]\centering
\scriptsize
\begin{adjustbox}{width=\linewidth}
    \begin{tabular}{p{1.45cm}|C{0.35cm}C{0.43cm}|C{0.35cm}C{0.43cm}|C{0.35cm}C{0.43cm}}
        \toprule
        \textbf{Method} & \textbf{PQ$\uparrow$}  & \textbf{\ospapsbold$\downarrow$}   & \textbf{PQ$^{\textbf{Th}}$$\uparrow$} & \textbf{\ospapsboldt$\downarrow$}  & \textbf{PQ$^{\textbf{St}}$$\uparrow$} &\textbf{\ospapsbolds $\downarrow$}\\
        \midrule
       \textbf{ODISE-L}\cite{xu2023open} & 10.57 & \textbf{0.85}  & 7.03 & 0.90 & \textbf{29.87} & \textbf{0.72}\\
        \textbf{ODISE-C}\cite{xu2023open} & \textbf{11.07} & \textbf{0.85} & \textbf{8.41} & \textbf{0.88} & 25.55 & 0.78 \\
        \textbf{FC-CLIP}\cite{yu2023convolutions} & 10.06 & 0.87 & 7.07 & 0.90 & 26.36 & 0.78\\
        \bottomrule
    \end{tabular}
\end{adjustbox}
\caption{\small\textbf{Results of SOTA OW panoptic segmentation models on JRDB-Panotrack testing set.} All models were trained solely on the COCO panoptic dataset and underwent zero-shot evaluation on JRDB. ODISE-L and ODISE-C represent the model with class label and caption label supervisions, respectively.}
\label{tab:panoptic:openworld_16th_11}
\end{table}

In JRDB-Panotrack, OSPA is preferred as it is an actual metric in mathematical terms, fulfilling the triangle inequality, not threshold-based, and treats masks equally regardless of their size, without penalising error rectification.

\section{Experiments}

To explore the distinct challenges of JRDB-PanoTrack, we first evaluate advanced panoptic segmentation in both 2D closed-world (CW) and open-world (OW) settings. Then we investigate panoptic tracking methods. We also briefly evaluate 3D CW segmentation and tracking using pseudo labels generated from 2D annotations. Note that all of the experiments presented are done using individual views, not stitched views. The results show that the JRDB-PanoTrack dataset provides a uniquely challenging environment for panoptic segmentation and tracking.

\noindent\textbf{Evaluation protocol} Due to the absence of pretrained models for close/open-world panoptic segmentation/tracking limits our evaluation in multi-label settings. We preprocess these areas into single-label ones, selecting \textit{thing} objects and omitting \textit{stuff} behind them, to utilize standard evaluation metrics alongside OSPA$_{PS}$ and OSPA$^{2}_{PT}$. We hope future research can exploit the full potential of our dataset's multi-class segmentation annotations.

\subsection{Panoptic Segmentation}
\textbf{Implementation.} We adopt the ResNet-50 backbone and COCO pertaining for all models, training with a batch size of 6 and a learning rate of $1 \times 10^{-4}$
over 110K iterations on 2 RTX4090 GPUs. For other settings, we adhere to the default configuration in \cite{cheng2021mask2former,li2023mask,yu2022k}. 
In OW experiments, we follow the official implementations of ODISE \cite{xu2023open} and FC-CLIP \cite{yu2023convolutions}. For all cross-domain experiments, we use the weights pretrained on COCO and infer on JRDB-PanoTrack. For in-domain experiments, we train FC-CLIP on our OW training set and infer on our OW test set. The model is trained with two RTX4090 GPUs with batch size 8, learning rate $5 \times 10^{-4}$, other training setting use same as \cite{yu2023convolutions}. We do not train ODISE due to its very high computational costs. 


\begin{table}[t]\centering
\scriptsize
    \begin{adjustbox}{width=\linewidth,center}
        \begin{tabular}{{C{0.6cm}|C{0.5cm}|C{0.3cm}C{0.43cm}|C{0.35cm}C{0.43cm}|C{0.35cm}C{0.4cm}}}
            \toprule
           \multicolumn{2}{c|}{\textbf{Train strategy}} &
            \multicolumn{2}{c|}{\textbf{All}} &
            \multicolumn{2}{c|}{\textbf{Thing}}&
            \multicolumn{2}{c}{\textbf{Stuff}}\\
            \cmidrule(l{0pt}r{0pt}){1-8} 
            \textbf{COCO} & \textbf{JRDB}& \textbf{PQ$\uparrow$}     & \ospapsbold $\downarrow$  &\textbf{PQ$^{\textbf{Th}}\uparrow$}& \ospapsboldt$\downarrow$   & \textbf{PQ$^{\textbf{St}}$$\uparrow$}&\ospapsbolds$\downarrow$  \\\midrule
           &  \textbf{\checkmark}  & 31.41& 0.67& 27.12  & 0.72 &43.88 & 0.53\\
          \textbf{\checkmark}&\textbf{\checkmark}&\textbf{36.57}&\textbf{0.64} & \textbf{33.07}&\textbf{0.68}&\textbf{46.74}&\textbf{0.51}\\
            \bottomrule
         \end{tabular}
    \end{adjustbox}
    \caption{{\small\textbf{CW panoptic segmentation results of MaskDino with different training strategies on JRDB-PanoTrack.} Top: we solely train the model on JRDB-PanoTrack. Bottom: we use COCO pertaining followed by finetuning on JRDB-PanoTrack.}}
    \label{tab:panoptic:ablation}
\end{table}

\begin{table}[t]
\scriptsize
\centering
\begin{adjustbox}{width=\linewidth}
    \begin{tabular}{C{0.7cm}|p{1.7cm}|C{0.32cm}C{0.43cm}|C{0.2cm}C{0.43cm}}
    \toprule
    \multirow{2}{*}{\textbf{Domain}} & 
    \multirow{2}{*}{\textbf{Method}} & 
    \multicolumn{2}{c|}{\textbf{Known}} & \multicolumn{2}{c}{\textbf{Unknown}}\\
    \cmidrule(l{0pt}r{0pt}){3-6}
    & & \textbf{PQ$\uparrow$} & \textbf{\ospapsbold$\downarrow$} & \textbf{PQ$\uparrow$} & \textbf{\ospapsbold$\downarrow$}\\
    \midrule
    \multirow{3}{*}{\textbf{Cross}}
        & \textbf{ODISE-L}\cite{xu2023open} & 14.94 & 0.84 & 3.86 & \textbf{0.92}\\
        & \textbf{ODISE-C}\cite{xu2023open} & 13.58 & 0.84 & \textbf{7.18} & \textbf{0.92}\\
        & \textbf{FC-CLIP}\cite{yu2023convolutions} & 14.29 & 0.86 & 3.56 & 0.93\\
    \midrule
    \textbf{In} & \textbf{FC-CLIP}\cite{yu2023convolutions} & \textbf{24.95} & \textbf{0.75} & 3.19 & 0.98\\
    \bottomrule
    \end{tabular}
\end{adjustbox}
\caption{\small\textbf{Performance of SOTA OW panoptic segmentation models on JRDB-PanoTrack.} Cross-domain methods are trained on COCO and tested on our dataset, while in-domain methods are trained with JRDB-PanoTrack \textit{known} classes. ODISE-L and ODISE-C represent the model with class and caption supervisions, respectively.}
\label{tab:panopticSeg_known_unknow_cross_indomain}
\end{table}


\noindent\textbf{CW Panoptic Segmentation.}
We evaluate SOTA methods on JRDB-PanoTrack (\cref{tab:panoptic:closeworld_16th_11}) and obtain the following findings: 
\textbf{(1)} Lower performance across all methods compared to COCO results, particularly for \textit{Thing} classes. 
This highlights challenges like complex \textit{Thing} instances and varied object scales in diverse environments.
\textbf{(2)} Mask DINO stands out, which achieves PQ of 36.57\% with COCO pertaining and 31.41\% without it (\cref{tab:panoptic:ablation}). One reason is that our dataset contains crowded objects, and Mask DINO contains more object queries to capture a mass of object candidates.
Meanwhile, MaskDino pretrained on JRDB-PanoTrack achieves higher performance on the COCO dataset (see the supplemental material), suggesting JRDB-PanoTrack's ability to generalize to other domains. These insights emphasize JRDB-PanoTrack's unique challenges in CW panoptic segmentation, leading us to further explore its role in OW panoptic segmentation setting.

\begin{table}[!t]
\scriptsize
\centering
\begin{adjustbox}{width=\columnwidth, center}
    \begin{tabular}{c|p{0.35cm}|p{0.3cm}p{0.3cm}p{0.3cm}p{0.3cm}p{0.3cm}p{0.5cm}}
        \toprule
        & \textbf{Trkr} & \textbf{STQ$\uparrow$} & \textbf{Frag$\downarrow$} & \textbf{IDF1$\uparrow$} & \textbf{\ospaptbold $\downarrow$} & \textbf{\ospaptboldt $\uparrow$} & \textbf{\ospaptbolds $\uparrow$} \\
        \midrule
        \multirow{3}{*}{\rotatebox[origin=c]{90}{\textbf{Kmax}}}      
                                                                      & \textbf{OS}   & 7.40  & \textbf{4239} & 28.70 & \textbf{0.805} & \textbf{0.867} & \textbf{0.546} \\
                                                                      & \textbf{BT}   & 8.77  & 6962           & 28.78 & 0.816 & 0.885 & 0.546 \\
                                                                      & \textbf{BS}   & \textbf{14.20} & 8932 & \textbf{29.76} & 0.831 & 0.909 & 0.546 \\
        \midrule
        \multirow{3}{*}{\rotatebox[origin=c]{90}{\textbf{2Former}}} 
                                                                      & \textbf{OS}   & 7.09  & \textbf{4162} & 27.49 & \textbf{0.799} & \textbf{0.863} & \textbf{0.530} \\
                                                                      & \textbf{BT}   & 8.58  & 7251           & 27.67 & 0.808 & 0.880 & 0.530 \\
                                                                      & \textbf{BS}   & \textbf{13.89} & 9392 & \textbf{29.33} & 0.817 & 0.902 & 0.530 \\
        \midrule
        \multirow{3}{*}{\rotatebox[origin=c]{90}{\textbf{DINO}}}       
                                                                      & \textbf{OS}   & 7.70  & \textbf{4515} & 30.40 & \textbf{0.793} & \textbf{0.851} & \textbf{0.548} \\
                                                                      & \textbf{BT}   & 9.00  & 7550           & 30.39 & 0.804 & 0.870 & 0.548 \\
                                                                      & \textbf{BS}   & \textbf{14.50} & 9609 & \textbf{31.61} & 0.822 & 0.901 & 0.548 \\
        \bottomrule
    \end{tabular}
\end{adjustbox}
\scriptsize\caption{\textbf{2D CW panoptic tracking results.} Kmax for Kmax-Deeplab, 2Former for Mask2Former, DINO for mask DINO, OS for OC-SORT, BT for ByteTrack, and BS for BoT-SORT. \ospaptboldt \text{ and } \ospaptbolds \text{ are} the OSPA$^2$ metric for \textit{Thing} and  \textit{Stuff} classes.}
\label{tab:trackers_evaluation_CW}
\end{table}

\noindent\textbf{OW Panoptic Segmentation.}
SOTA OW panoptic segmentation methods like FC-CLIP \cite{yu2023convolutions} and ODISE \cite{xu2023open} 
show notably lower performance on JRDB-PanoTrack (\cref{tab:panoptic:openworld_16th_11}) compared to other datasets. For instance, the PQ of FC-CLIP on ADE20K is $26.8$ while $10.06$ in our dataset, highlighting JRDB-PanoTrack's distinct and challenging nature, especially in recognizing and segmenting \textit{Unknown} classes. \cref{tab:panopticSeg_known_unknow_cross_indomain} further shows cross- 
\noindent and in-domain evaluations for \textit{Known} and \textit{Unknown} classes on our dataset. Cross-domain results indicate that while prior knowledge from other datasets like COCO aids in understanding \textit{Known} classes, it falls short with \textit{Unknown} classes, underlining JRDB-PanoTrack's OW segmentation challenge. In contrast, in-domain training improves the segmentation performance for \textit{Known} classes, but slightly impacts \textit{Unknown} classes, suggesting that new approaches are needed to address this core challenge. Additionally, we assess the transferability of JRDB-PanoTrack knowledge to other domains. Training exclusively on JRDB-PanoTrack yields a $13.7$ PQ on COCO (\cref{tab:open_world_cross_domain}), demonstrating effective knowledge transfer to different domains. This finding indicates the potential usage of JRDB-PanoTrack to improve segmentation performance on other domains.

\begin{table}[!t]\centering
\scriptsize
\renewcommand{\arraystretch}{0.8}
\begin{adjustbox}{width=\linewidth,center}
    \begin{tabular}{ll|c|c|c}
        \toprule
        \textbf{JRDB} & \textbf{COCO} & \textbf{PQ$\uparrow$} & \textbf{PQ$^{Th}$ $\uparrow$} & \textbf{PQ$^{St}$ $\uparrow$}\\\midrule
        & \textbf{\checkmark} & 42.16 & 47.43 & 34.20\\
        \textbf{\checkmark} & \textbf{\checkmark} & \textbf{44.48} & \textbf{50.48} & \textbf{35.43}\\
        \bottomrule
    \end{tabular}
\end{adjustbox}
\caption{CW panoptic segmentation results on COCO val of Mask DINO\cite{li2023mask} using different training data. Top: we solely train the model on COCO. Bottom: we use JRDB-PanoTrack pertaining followed by finetuning on COCO. JRDB refers to the JRDB-PanoTrack.}
\label{tab:closed_world_onCOCO}
\end{table}
\begin{table}[!t]
    \centering
    \begin{adjustbox}{width=\columnwidth,center}
    \small
    \begin{tabular}{l|C{11mm}C{11mm}C{11mm}C{11mm}}
        \toprule
        \textbf{Method} & \ospapsbold$\downarrow$& $\bm{\mathcal{O}}^{\textbf{Small}}_{\textbf{PS}}\downarrow$& $\bm{\mathcal{O}}^{\textbf{Medium}}_{\textbf{PS}}\downarrow$ & $\bm{\mathcal{O}}^{\textbf{Large}}_{\textbf{PS}}\downarrow$ \\
        \midrule
        \textbf{Kmax\cite{yu2022k}} & 0.670 & 0.823 & 0.596 & 0.370 \\
        \textbf{Mask2Former\cite{cheng2021mask2former}} & 0.655 & 0.805 & 0.589 & 0.371 \\
        \textbf{MaskDINO\cite{li2023mask}  } & \textbf{0.636} & \textbf{0.785} & \textbf{0.552} & \textbf{0.364} \\
        \bottomrule
    \end{tabular}
    \end{adjustbox}
    \caption{CW panoptic segmentation results on JRDB-PanoTrack testing for objects of different scales.}
    \label{tab:cw_ospa_per_scale}
\end{table}

\begin{table}[!t]
    \centering
    \begin{adjustbox}{width=\columnwidth,center}
    \small
    \begin{tabular}{l|C{11mm}C{11mm}C{11mm}C{11mm}}
        \toprule
        \textbf{Method} & \ospapsbold$\downarrow$& $\bm{\mathcal{O}}^{\textbf{Small}}_{\textbf{PS}}\downarrow$& $\bm{\mathcal{O}}^{\textbf{Medium}}_{\textbf{PS}}\downarrow$ & $\bm{\mathcal{O}}^{\textbf{Large}}_{\textbf{PS}}\downarrow$ \\
        \midrule
        \textbf{ODISE-L\cite{xu2023open}} & 0.851 & 0.950 & 0.790 & 0.550 \\
        \textbf{ODISE-C\cite{xu2023open}} & 0.849 & 0.906 & 0.696 & 0.431 \\
        \textbf{FC\_CLIP\cite{yu2023convolutions}} & 0.868 & 0.968 & 0.863 & 0.619 \\
        \textbf{FC\_CLIP+\cite{yu2023convolutions}} & \textbf{0.776} & \textbf{0.877} & \textbf{0.597} & \textbf{0.390} \\
        \bottomrule
    \end{tabular}
    \end{adjustbox}
    \caption{OW panoptic segmentation results on JRDB-PanoTrack testing for objects of different scales.}
    \label{tab:ow_ospa_per_scale}
\end{table}

\noindent\textbf{\textbf{Generalizability of JRDB-Panotrack}}
\cref{tab:closed_world_onCOCO} presents comparative results of the Mask DINO model on the COCO validation set, highlighting the generalizability of JRDB-Panotrack. 
Specifically, \cref{tab:closed_world_onCOCO} compares the performance when trained solely with the COCO dataset against a combined training scheme that includes both JRDB-PanoTrack and COCO datasets. Notably, the model pretrained on JRDB-PanoTrack followed by COCO tuning shows superior performance across all metrics compared to the model trained on COCO alone, which supporting JRDB-Panotrack is also beneficial for other domains.

\noindent\textbf{Knowledge transfer} \cref{tab:closed_world_onCOCO} demonstrates knowledge transferability between 
\begin{wraptable}{r}{0.45\linewidth} 
\centering
\small
\begin{tabular}{cc}
    \toprule
    \textbf{Train data} & \textbf{PQ} \\
    \midrule
    \textbf{COCO}  & 10.06 \\
    \textbf{JRDB} & 13.70 \\
    \bottomrule
\end{tabular}
\caption{\small\textbf{Cross-dataset validation results of FC-CLIP.} The first row indicates training on COCO and testing on JRDB-PanoTrack, while the second row is the opposite.}
\label{tab:open_world_cross_domain}
\end{wraptable}
datasets in open-world setting. In the closed-world setting, we also show that the knowledge from JRDB-PanoTrack can help improve performance when fine-tuning on the COCO dataset (\cref{tab:open_world_cross_domain}) and vice versa (\cref{tab:panoptic:ablation}). 
The results in both open-world and closed-world settings show that using JRDB-PanoTrack improves segmentation performance in other domains. This suggests that the size of JRDB-PanoTrack does not significantly hinder the performance.

\subsection{Panoptic Tracking} \label{sec: PT}
\textbf{Implementation.} 
We utilize default settings and implementations of recent popular tracking algorithms: ByteTrack\cite{bytetrack}, OC-SORT\cite{ocsort} and BoT-SORT\cite{botsort}. Masks predicted from CW and OW segmentation models are converted into bounding boxes and then fed into these trackers. 

\noindent\textbf{CW Panoptic Tracking.} In \cref{tab:trackers_evaluation_CW}, our evaluation highlights diverse capabilities of SOTA tracking methods. BoT-SORT excels in STQ and IDF1 metrics, showcasing its proficiency in object tracking and identity maintenance. OC-SORT, favored by the \ospapt\ metric, excels in consistently identifying objects across frames while minimizing noisy tracklets. BoT-SORT's performance, though strong in tracking objects, shows signs of instability, often losing track and struggling with consistent ID maintenance. Breaking down into \textit{Thing} and \textit{Stuff} classes, \ospapts \text{ error} for \textit{Stuff} remains constant because cardinality error is not penalised. The lower performance on our JRDB-Panotrack, compared to other datasets, can be attributed to our dense annotations and numerous tracklets, posing a significant challenge for segmentation and tracking.

\begin{table}[t]
    \scriptsize
    \centering
    \begin{adjustbox}{width=\columnwidth, center}
        \begin{tabular}{c|p{0.35cm}|p{0.3cm}p{0.3cm}p{0.3cm}p{0.3cm}p{0.3cm}p{0.5cm}}
            \toprule
            & \textbf{Trkr} & \textbf{STQ$\uparrow$} & \textbf{Frag$\downarrow$} & \textbf{IDF1$\uparrow$}  & \textbf{\ospaptbold $\downarrow$} & \textbf{\ospaptboldk $\downarrow$} & \textbf{\ospaptboldu $\downarrow$} \\
            \midrule
            \multirow{3}{*}{\rotatebox[origin=c]{90}{\smaller{FC\_CLIP}}}  & \textbf{OS}   & 2.50                    & \textbf{833} & 8.43 & \textbf{0.910} & \textbf{0.861} & \textbf{0.962} \\
            & \textbf{BT}   & 2.78                    & 1126                       & 8.82                     & 0.921                             & 0.871                              & 0.971                              \\
            & \textbf{BS}   & \textbf{4.71}           & 1956                       & \textbf{9.52}            & 0.921                             & 0.869                              & 0.979                              \\
            \midrule
            \multirow{3}{*}{\rotatebox[origin=c]{90}{\smaller{ODISE-L}}}   & \textbf{OS}   & 3.11                    & \textbf{1373} & 8.89 & \textbf{0.924} & \textbf{0.854} & \textbf{0.977} \\
            & \textbf{BT}   & 3.71                    & 2013                       & 9.31                     & 0.928                             & 0.867                              & 0.979                              \\
            & \textbf{BS}   & \textbf{6.49}           & 3167                       & \textbf{9.97}            & 0.927                             & 0.873                              & 0.980                              \\
            \midrule
            \multirow{3}{*}{\rotatebox[origin=c]{90}{\smaller{ODISE-C}}}   & \textbf{OS}   & 4.19                    & \textbf{1112} & 9.22 & \textbf{0.917} & \textbf{0.862} & 0.979 \\
            & \textbf{BT}   & 5.07                    & 1457                       & 9.34                     & 0.925                             & 0.866                              & \textbf{0.978}                              \\
            & \textbf{BS}   & \textbf{8.32}           & 2139                       & \textbf{10.80}           & 0.924                             & 0.863                              & 0.980                              \\
            \midrule
            \multirow{3}{*}{\rotatebox[origin=c]{90}{\smaller{FC\_CLIP+}}} & \textbf{OS}   & 4.90                    & \textbf{2833} & 13.20 & \textbf{0.897} & \textbf{0.826} & \textbf{0.990} \\
            & \textbf{BT}   & 5.53                    & 4614                       & 13.44                    & 0.905                             & 0.836                              & 0.993                              \\
            & \textbf{BS}   & \textbf{9.01}           & 6868                       & \textbf{14.51}           & 0.908                             & 0.850                              & 0.993                              \\
            \bottomrule
        \end{tabular}
    \end{adjustbox}
    \scriptsize\caption{ \textbf{2D OW panoptic tracking results.} (OS for OC-SORT, BT for ByteTrack and BS for BoT-SORT). \ospaptboldk \text{ and } \ospaptboldu \text{ are} the OSPA$^2$ metric for \textit{Known} and  \textit{Unknown} classes. ODISE-L and ODISE-C represent ODISE using class names and captions as supervison.}
    \label{tab:trackers_evaluation_OW}
\end{table}

\noindent\textbf{OW Panoptic Tracking.} OW panoptic tracking results, as shown in \cref{tab:trackers_evaluation_OW}, indicate a different set of challenges. While BoT-SORT is good at maintaining object identities and delivering high-quality segmentation, it exhibits higher fragmentation, indicating inconsistency in track identity over time. In contrast, OC-SORT, though it may not always top the STQ or IDF1 scores, shows greater consistency with fewer fragmentations and lower OSPA errors. The overall lower performance on the JRDB-Panotrack dataset reflects the complexities of OW tracking, especially when handling \textit{unknown} objects. This underscores the need for advanced tracking algorithms to adapt to unfamiliar objects and maintain consistent track identities.

\subsection{3D Panoptic Segmentation \& Tracking}

\begin{table}[!tp]
    \centering
    \small
    \begin{adjustbox}{width=\linewidth,center}
        \begin{tabular}{p{17mm}|C{7mm}C{7mm}C{7mm}C{7mm}C{7mm}C{7mm}}
            \toprule
            \multicolumn{7}{c}{\textbf{3D panoptic segmentation}} \\\midrule
            \textbf{Method} & \textbf{IoU}$\uparrow$ & \textbf{PQ}$\uparrow$ & \textbf{RQ}$\uparrow$ & \ospaptbold$\downarrow$ & \ospaptboldc$\downarrow$ & \ospaptboldl$\downarrow$ \\\cmidrule{1-7}
            \textbf{DSNet\cite{dsnet}}   & 12.62          & 3.41                 & 4.25               & 0.843          & 0.657          & 0.186          \\
            \textbf{Mask4D\cite{mask4d}}  & 13.51          & 3.57                 & 5.39               & 0.826          & 0.643          & 0.183          \\
            \textbf{MaskPLS\cite{maskpls}} & \textbf{15.13} & \textbf{7.02}        & \textbf{10.74}     & \textbf{0.795} & \textbf{0.629} & \textbf{0.166}\\\midrule
            \multicolumn{7}{c}{\textbf{3D Panoptic Tracking}} \\\midrule
           \textbf{Method}  & \textbf{LSTQ}$\uparrow$  & \textbf{S}$_{assoc}$$\uparrow$ & $\textbf{S}_{cls}$$\uparrow$ & \ospaptbold$\downarrow$    & \ospaptboldc$\downarrow$ & \ospaptboldl$\downarrow$ \\\cmidrule{1-7}
            \textbf{DSNet\cite{dsnet}}   & 25.35          & 55.18                & 11.64              & 0.882          & 0.726          & 0.156          \\
            \textbf{Mask4D\cite{mask4d}}  & \textbf{27.87} & \textbf{66.32}       & \textbf{11.71}     & \textbf{0.860} & \textbf{0.711} & \textbf{0.149} \\
            \bottomrule
        \end{tabular}
    \end{adjustbox}
    \caption{Results for 3D panoptic segmentation and tracking on JRDB-PanoTrack testing. $S_{assoc}$ and $S_{cls}$ are association and classification scores (components of LSTQ), respectively. \ospaptboldc and \ospaptboldl are OSPA cardinality and localisation errors (components of \ospaptbold), as explained in \cite{ospa}.}
    \label{tab:3Dclosedworld}
\end{table}

In this work, we briefly touch on 3D CW panoptic segmentation and tracking, though it is not the main focus of this paper. Specifically, we projected 2D panoptic labels onto 3D point clouds, using these projections as pseudo-labels for model training.
For evaluation, we use our proposed OSPA, OSPA$^2$ and adopt popular metrics for 3D panoptic segmentation (PQ, IOU) and tracking (LSTQ\cite{aygun20214d}). It's important to note that these 3D pseudo-labels may contain noise, potentially affecting result accuracy. In terms of 3D panoptic segmentation, as shown in \cref{tab:3Dclosedworld}, MaskPLS emerges as the superior method, excelling in all metrics. This indicates MaskPLS's enhanced ability to identify and segment objects precisely in 3D space. In 3D panoptic tracking, Mask4D takes the lead in LSTQ\cite{aygun20214d} and achieves the best OSPA score of 0.860, denoting its strength in maintaining object identities and tracking consistency over time. Also, the higher $S_{assoc}$ scores compared to $S_{cls}$, suggesting that these methods are better at object association and tracking than at precise classification in a 3D environment.


\section{Conclusion}
In this paper, we have introduced the \textit{JRDB-PanoTrack} dataset, a novel dataset designed for open-world panoptic segmentation and tracking, particularly for robotics and vision applications. The uniqueness and complexity of JRDB-PanoTrack set it apart from the existing ones. Our extensive evaluations underscore the dataset's challenges, emphasizing the necessity for more robust methodologies in both closed-world and open-world scenarios. The dataset offers new ground for future research, especially in developing algorithms that can effectively handle densely populated environments and diverse object interactions that are typical in real-world settings.\\


\noindent{\textbf{Acknowledgments.} This work has been partially funded by The Australian Research Council Discovery Project (ARC DP2020102427). We also acknowledge the partial sponsorship of our research by the DARPA Assured Neuro Symbolic Learning and Reasoning (ANSR) program, under award number FA8750-23-2-1016.} 

{
    \small
    \bibliographystyle{ieeenat_fullname}
    \bibliography{main}
}

\end{document}